\documentclass[journal]{IEEEtran}
\usepackage{cite}

\usepackage{amsfonts}
\usepackage{array}
\usepackage{url}
\usepackage{amsmath}
\usepackage{amssymb}
\usepackage{color}

\ifCLASSINFOpdf
   \usepackage[pdftex]{graphicx}
   \graphicspath{{../pdf/}{../jpeg/}}
   \DeclareGraphicsExtensions{.pdf,.jpeg,.png}
\else
   \usepackage[dvips]{graphicx}
   \graphicspath{{../eps/}}
   \DeclareGraphicsExtensions{.eps}
\fi

\ifCLASSOPTIONcompsoc
  \usepackage[caption=false,font=normalsize,labelfont=sf,textfont=sf]{subfig}
\else
  \usepackage[caption=false,font=footnotesize]{subfig}
\fi

\begin{document}

\title{FreDFT: Frequency Domain Fusion Transformer for Visible-Infrared Object Detection}

\author{Wencong~Wu,~Xiuwei~Zhang,~Hanlin~Yin,~Shun~Dai,~Hongxi~Zhang,~Yanning~Zhang,~\IEEEmembership{Fellow,~IEEE}
\thanks{This research is supported in part by the National Natural Science Foundation of China (62476220), National Key Research and Development Program of China (2023YFC3209305, 2023YFC3209304), Guangdong Basic and Applied Basic Research Foundation (2024A1515030186), and Natural Science Basic Research Program of Shaanxi (2024JC-DXWT-07, 2024JC-YBQN-0719). (Corresponding authors: Xiuwei Zhang; Hanlin Yin.)}
\thanks{Wencong Wu, Shun Dai and Hongxi Zhang are with the School of Computer Science, Northwestern Polytechnical University, Xi'an 710072, China (Email: wencongwu@mail.nwpu.edu.cn; shundai@mail.nwpu.edu.cn; hongxi0209@mail.nwpu.edu.cn).}
\thanks{Xiuwei Zhang, Hanlin Yin and Yanning Zhang are with the Shaanxi Provincial Key Laboratory of Speech and Image Information Processing and the National Engineering Laboratory for Integrated Aerospace-Ground-Ocean Big Data Application Technology, School of Computer Science, Northwestern Polytechnical University, Xi'an 710072, China (Email: xwzhang@nwpu.edu.cn; iverlon1987@nwpu.edu.cn; ynzhang@nwpu.edu.cn).}
}

\markboth{Manuscript submitted to XXX}%
{Shell \MakeLowercase{\textit{ et al.}}: Bare Demo of IEEEtran.cls for IEEE Journals Journal of \LaTeX\ Class Files,~Vol.~14, No.~8, August~2015}
\maketitle

%the initial submission can not exceed 10 pages.

\begin{abstract}
Visible-infrared object detection has gained sufficient attention due to its detection performance in low light, fog, and rain conditions. However, visible and infrared modalities captured by different sensors exist the information imbalance problem in complex scenarios, which can cause inadequate cross-modal fusion, resulting in degraded detection performance. Furthermore, most existing methods use transformers in the spatial domain to capture complementary features, ignoring the advantages of developing frequency domain transformers to mine complementary information. To solve these weaknesses, we propose a frequency domain fusion transformer, called FreDFT, for visible-infrared object detection. The proposed approach employs a novel multimodal frequency domain attention (MFDA) to mine complementary information between modalities and a frequency domain feed-forward layer (FDFFL) via a mixed-scale frequency feature fusion strategy is designed to better enhance multimodal features. To eliminate the imbalance of multimodal information, a cross-modal global modeling module (CGMM) is constructed to perform pixel-wise inter-modal feature interaction in a spatial and channel manner. Moreover, a local feature enhancement module (LFEM) is developed to strengthen multimodal local feature representation and promote multimodal feature fusion by using various convolution layers and applying a channel shuffle. Extensive experimental results have verified that our proposed FreDFT achieves excellent performance on multiple public datasets compared with other state-of-the-art methods. The code of our FreDFT is linked at https://github.com/WenCongWu/FreDFT.
\end{abstract}

\begin{IEEEkeywords}
Visible-infrared object detection, Transformer, Multimodal frequency domain attention, Feature fusion, Modality imbalance.
\end{IEEEkeywords}

\IEEEpeerreviewmaketitle

\section{INTRODUCTION}
\IEEEPARstart{O}{bject} detection, as a fundamental and significant task in the field of computer vision, aims to identify the location of objects in images and determine their categories. Many previous studies have focused on visible light (RGB) images with rich textures and colors, such as Faster R-CNN \cite{Ren17} and YOLO \cite{YOLOv5, YOLOv8}. However, RGB images taken at night or in foggy and rainy environments have poor quality. Infrared (IR) images utilize the thermal radiation emitted by the object itself to form an image, which is not affected by the above interference conditions. It can be concluded that the complementarity between RGB and IR images can provide important semantic information for multispectral object detection and has the potential to improve detection performance \cite{LiZ23}, \cite{Li24}, \cite{Hu25}.

However, there is a feature imbalance problem between modalities. Since RGB and IR modalities are captured by different sensors, their different imaging ways lead to semantic inconsistency of cross-modal features. Previous many methods \cite{Fang21}, \cite{Cao23} have not considered this problem, resulting in information conflict and low-quality fusion results during cross-modal feature fusion. Specifically, Fang et al. \cite{Fang21} utilized a dual backbone composed of two CSPDarknet53 networks to extract RGB and IR features, fed these merged multimodal features obtained by the concatenation operation into the designed cross-modality fusion transformer (CFT) to generate the final fusion results. Cao et al. \cite{Cao23} employed two-branch ResNet \cite{He16} to extract multimodal features from RGB and IR images, and a channel switching and spatial attention (CSSA) was designed to exchange channel features and enhance spatial information between RGB and IR modalities to better fuse multispectral features in spatial-channel levels. Although the above multimodal methods have achieved better detection performance than the monomodal methods \cite{Ren17, YOLOv5, YOLOv8}, they overlook the existing feature imbalance problem, which interferes with the multispectral detection performance. To this end, Zhou et al. \cite{Zhou20} first focused on the modality feature imbalance problem in depth by designing a differential modality aware fusion module to obtain complementary multimodal features, but only considered cross-modal channel information interaction and neglected spatial information interaction, resulting in the inadequate solution of intermodal feature imbalance problem.

In addition, most existing works tend to use spatial domain transformers to capture complementary information across modalities to facilitate feature fusion, but neglect the utilization of frequency domain transformers. For example, Xie et al. \cite{Xie23} developed a feature interaction and self-attention fusion network to promote multimodal information exchange and enhancement to obtain informative fusion features. Yuan et al. \cite{Yuan24} designed a calibrated and complementary transformer in the spatial domain for remote sensing RGB-IR object detection, where an inter-modality cross-attention is developed to generate the calibrated and complementary features for modality fusion. Although these methods have achieved great detection results, they all perform feature enhancement and fusion in the spatial domain and ignore the importance of frequency domain information. Moreover, due to the noise interference and object occlusion of RGB and IR images, multimodal feature representation and interaction only in the spatial domain limits the discrimination ability of the detection model \cite{Weng24}. In the frequency domain, the texture details of RGB modality and the structural thermal features of IR modality can be more fundamentally decoupled and fused, thereby achieving more robust and accurate detection in complex scenarios. Zeng et al. \cite{Zeng24} firstly perform multimodal feature enhancement in the frequency domain for RGB-IR object detection, and then feed these enhanced features into spatial domain transformers to capture complementary information. In the field of RGB-IR object detection, existing studies have neglected to combine frequency domain and transformers to construct frequency domain transformers to extract complementary information, resulting in the inability to obtain robust multimodal feature representations in complex environments.

To address these issues, we propose a novel frequency domain fusion transformer, namely FreDFT, for visible-infrared object detection. The FreDFT is composed of a local feature enhancement module (LFEM), a cross-modal global modeling module (CGMM), and a frequency domain feature aggregation module (FDFAM). Specifically, we exploit the LFEM to enhance multimodal multi-scale local features from two backbone networks via different convolution layers and a channel shuffle, and apply the CGMM to adequately reduce intermodality differences at the spatial and channel levels through global interaction and representation of multimodal features. Then, the latent feature correlations between RGB and IR modalities are accurately captured by a multimodal frequency domain attention (MFDA) in the FDFAM, where we estimate the scaled dot product attention using element-wise product operations in the frequency domain instead of matrix multiplications in the spatial domain according to the convolution theorem. Furthermore, a frequency domain feed-forward layer (FDFFL) in the FDFAM is designed for exploiting multi-scale representations to promote multimodal feature combination by using a mixed-scale frequency feature fusion strategy. This proposed innovative approach can strengthen multimodal features and utilize the frequency domain transformer to optimize the performance of multispectral object detection.

In summary, the main contributions of this work are as follows.

(1) The proposed FreDFT can enhance local multimodal features, mitigate intermodality differences, and better fuse cross-modal features for visible-infrared object detection task, which outperforms other state-of-the-art methods on three public datasets, including the FLIR, LLVIP, and M$^3$FD.

(2) A novel FDFAM is designed to effectively merge multimodal features, which consists of a MFDA and a FDFFL. The MFDA can capture the correlations between different modalities to facilitate cross-modal feature fusion, and the FDFFL integrates multi-scale frequency information into the global representation.

(3) To promote better fusion of multimodal features, a LFEM is developed by using different convolution kernels to focus on different local information locations and applying a channel shuffle to enhance the channel information interaction, and a new CGMM is introduced to alleviate multimodal heterogeneity by cross-modal pixel-wise spatial-channel interaction to explore the inherent relations of cross-modal features.

The rest of this work is arranged as follows. In Section \ref{Sec_Related_work}, we review the existing methods for visible-infrared object detection. Section \ref{Sec_Methodology} reports our proposed FreDFT. The experimental settings and comparisons are shown in Section \ref{Sec_Experiments}. Finally, we conclude our work in Section \ref{Sec_Conclusion}.

\section{RELATED WORK}\label{Sec_Related_work}
\subsection{Visible-infrared object detection}
Visible-infrared object detection has received much attention due to its excellent object discrimination ability in complex scenes. Zhang et al. \cite{Zhang21} introduced a guided attentive feature fusion (GAFF) method to effectively merge multimodal features extracted by two-branch ResNet18 \cite{He16} from RGB and IR images, and adopted the RetinaNet detector \cite{Lin17} to predict multispectral detection results. Yan et al \cite{Yan23} designed a cross-modality complementary information fusion network (CCIFNet) to execute non-local interactions between RGB and IR modalities and keep the spatial relationship of multimodalities for multispectral pedestrian detection. Cao et al. \cite{Cao23} used two ResNet50\cite{He16} as the backbone network to exploit RGB and IR features, and applied the channel switching and spatial attention (CSAA) to fuse different modalities in spatial and channel levels, where Faster R-CNN detector \cite{Ren17} were selected to generate the prediction results. Zhang et al. \cite{ZhangY23} proposed an effective Triple-I Net (TINet) for RGB-IR object detection by employing an illumination-guided feature weighting module to promote network learning, where multimodal features were acquired by using two same ResNet50 with the FPN \cite{LinD17}. A removal and selection detector (RSDet) \cite{Zhao24} was designed for RGB-IR object detection, where the redundant spectrum removal module was used to remove noise information in multimodalities. Moreover, A lightweight cross-modality attentive feature fusion (CMAFF) approach was presented by Fang et al. \cite{Fang22} to fuse multimodal features captured by the two-stream CSPDarknet53 in the YOLOv5 detector, and these fused features were sent to the detection head to obtain the detection results. Later, Xie et al. \cite{Xie23} developed a feature interaction and self-attention fusion network (FISAFN) for multispectral object detection, where the two-branch CSPDarknet53 and detection head of the YOLOv5 were used for extracting inter-modality features and producing the final results, respectively. These methods all perform feature fusion between modalities and then feed the fused features into the detector to obtain multispectral detection results. Liu et al. \cite{Liu22} designed a target-aware dual adversarial learning (TarDAL) network for infrared and visible image fusion, and then adopted the YOLOv5 model on the fused image to gain the detection results. Chen et al. \cite{Chen22} applied a late fusion strategy to integrate the detection results of different modalities by a probabilistic ensembling technique. Although these CNN-based multispectral detection methods have achieved great performance, they mainly focus on local information exchange and fusion between modalities, neglecting the mining of long-range dependencies of multimodal information, resulting in limited performance in complex scenes, especially in object localization under partial occlusion.

\begin{figure*}[htbp]
	\begin{center}
		\includegraphics[width=0.75\textwidth]{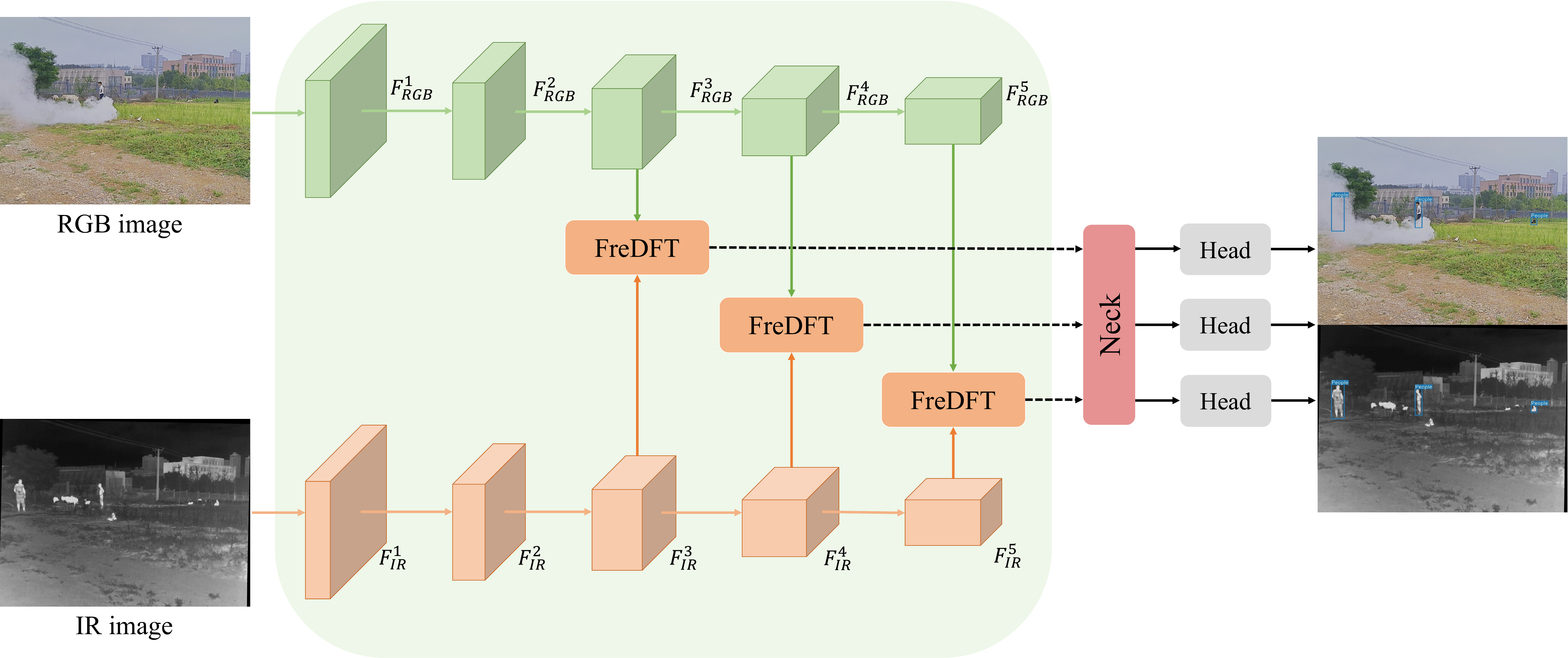}
		\caption{The architecture of the proposed RGB-IR object detection framework. The FreDFT stands for our proposed frequency domain fusion transformer, which is used to merge multimodal features from the dual backbone network effectively, and these fused features are fed into the neck and detection head to generate the prediction results.}
        \label{fig1:Framework}
	\end{center}
\end{figure*}

\subsection{Transformers in visible-infrared object detection}
To capture long-range relationships between different modalities and better fuse multimodal features, many transformer-based visible-infrared detection models have been developed. Fang et al. \cite{Fang21} designed a cross-modality fusion transformer (CFT) for RGB-IR object detection, where the CFT was embedded in two-branch CSPDarknet53 from the YOLOv5 backbone network to exploit long-range dependencies and fuse global contextual information, which makes it achieve competitive multispectral detection accuracy. You et al. \cite{You23} proposed a multi-scale aggregation network (MSANet) for multispectral object detection, where the multi-scale aggregation transformer using the multi-head self-attention scheme was designed to capture rich details and texture information from RGB and IR modalities, and the cross-modal merging fusion mechanism was applied for aggregating the complementary information from various modalities. To achieve great multimodal object detection, Dong et al. \cite{Dong24} introduced a dual transformer feature fusion module to merge local and global features from two modalities, and utilized a contrastive learning strategy to promote complementary mining and information interaction of multimodal features. Yang et al. \cite{Yang24} designed a multidimensional fusion network (MMFN) for RGB-IR object detection, where local, global, and channel multimodal features were exploited to better fuse complementary information of different modalities. Lee et al. \cite{Lee24} proposed a new multispectral object detector, namely CrossFormer, based on a hierarchical transformer and cross-guidance strategy, where the cross-guided attention module (CGAM) consists of two parallel transformers with the multi-head self-attention mechanism to achieve inter-modality information exchange. Shen et al. \cite{Shen24} presented a novel cross-attention guided feature fusion framework with an iterative learning strategy, called ICAFusion, for multispectral object detection, where the dual cross-attention transformer was introduced to capture cross-modal complementary information and perform global multimodal feature interaction. Despite these progresses, transformer-based cross-modal detection methods still have high computational complexity in using self-attention and cross-attention mechanisms to calculate the similarity between tokens in the spatial domain. To the best of our knowledge, this work is the first to design a multimodal frequency domain transformer for visible-infrared object detection task. Further, since RGB and IR images with different imaging patterns own discrepant image properties \cite{Zhou20}, many existing methods use dual backbone network to extract multimodal features and directly employ the transformer architecture to fuse inter-modality features, which does not eliminate the intermodal differences and may lead to the detection performance degradation.

\section{METHODOLOGY}\label{Sec_Methodology}
\subsection{Proposed architecture}
\begin{figure}[htbp]
	\begin{center}
		\includegraphics[width=0.40\textwidth]{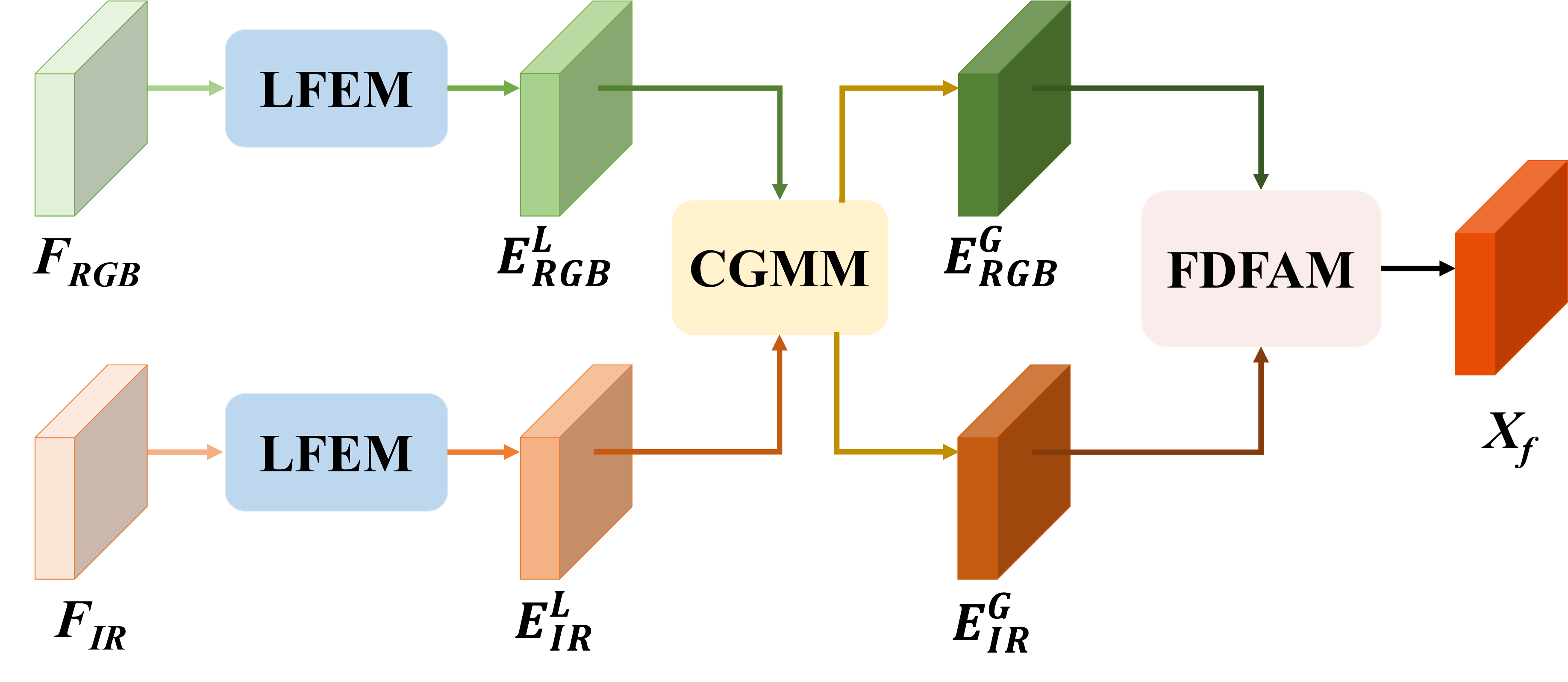}
		\caption{The structure of the proposed frequency domain fusion transformer (FreDFT). The LFEM, CGMM, and FDFAM denote the local feature enhancement module, cross-modal global modeling module, and frequency domain feature aggregation module, respectively. $E_{RGB}^L$ and $E_{IR}^L$ are the output of the LFEM, and $E_{RGB}^G$ and $E_{IR}^G$ are the output of the CGMM. $X_f$ is the fused feature.}
        \label{fig2:FreDFT}
	\end{center}
\end{figure}

Our designed visible-infrared object detection architecture is displayed in Fig. \ref{fig1:Framework}. Following the latest ICAFusion \cite{Shen24}, CrossFormer \cite{Lee24}, and MMFN \cite{Yang24}, our detection framework also selects the YOLOv5 \cite{YOLOv5} as the reference detector, which consists of a dual backbone network, three FreDFTs, a neck, and multi-scale detection head. The dual backbone network is two identical CSPDarknet53, which can extract multi-scale features from RGB and IR modalities. Our proposed frequency domain fusion transformer (FreDFT) is applied to fuse multi-scale cross-modal features. The structure of the FreDFT can be seen in Fig. \ref{fig2:FreDFT}, which consists of two local feature enhancement modules (LFEMs), a cross-modal global modeling module (CGMM), and a frequency domain feature aggregation module (FDFAM). The LFEM is constructed to enhance cross-modal local features, the CGMM is used to eliminate intermodal differences, and then the enhanced multimodal features are effectively merged by the FDFAM. It should be noted that the last three pairs of feature maps $F_{RGB}^i$ and $F_{IR}^i$ (i $\in$ \{3, 4, 5\}) are aggregated by our FreDFT. These high-dimensional features can meet the semantic information required for detection tasks and reduce computational complexity \cite{Fang22}.

\subsection{Local feature enhancement module}
\begin{figure}[htbp]
	\begin{center}
		\includegraphics[width=0.45\textwidth]{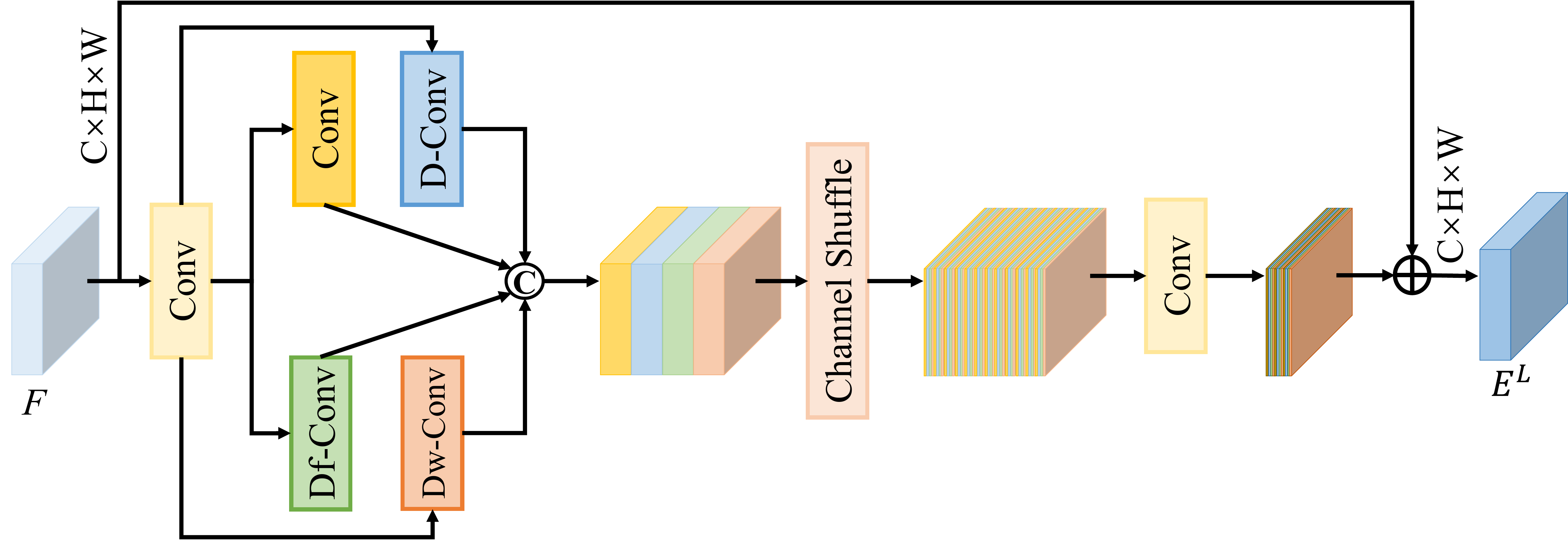}
		\caption{The structure of the designed local feature enhancement module (LFEM).}
        \label{fig3:LFEM}
	\end{center}
\end{figure}

Our proposed local feature enhancement module (LFEM) can be seen in Fig. \ref{fig3:LFEM}. Firstly, we adopt \text{$1\times1$} convolution layer with the batch normalization (BN) and SiLU activation function to increase nonlinearity without changing the size of the feature map. Further, a standard \text{$3\times3$} convolution layer (`Conv'), a dilated convolution layer (`D-Conv'), a deformable convolution layer (`Df-Conv'), and a depth-wise convolution layer (`Dw-Conv') are used for capturing feature information of different regions to enhance local information extraction, where different convolution layers are equipped with the BN and SiLU. Later,  we apply a concatenation operation to fuse the extracted information and rearrange the fused feature by a channel shuffle operation to improve model expression ability. Additionally, a \text{$1\times1$} convolution layer is used for feature dimensionality reduction. Finally, the initial information is added to the enhanced feature to augment the local feature.

\subsection{Cross-modal global modeling module}
\begin{figure}[htbp]
	\begin{center}
		\includegraphics[width=0.45\textwidth]{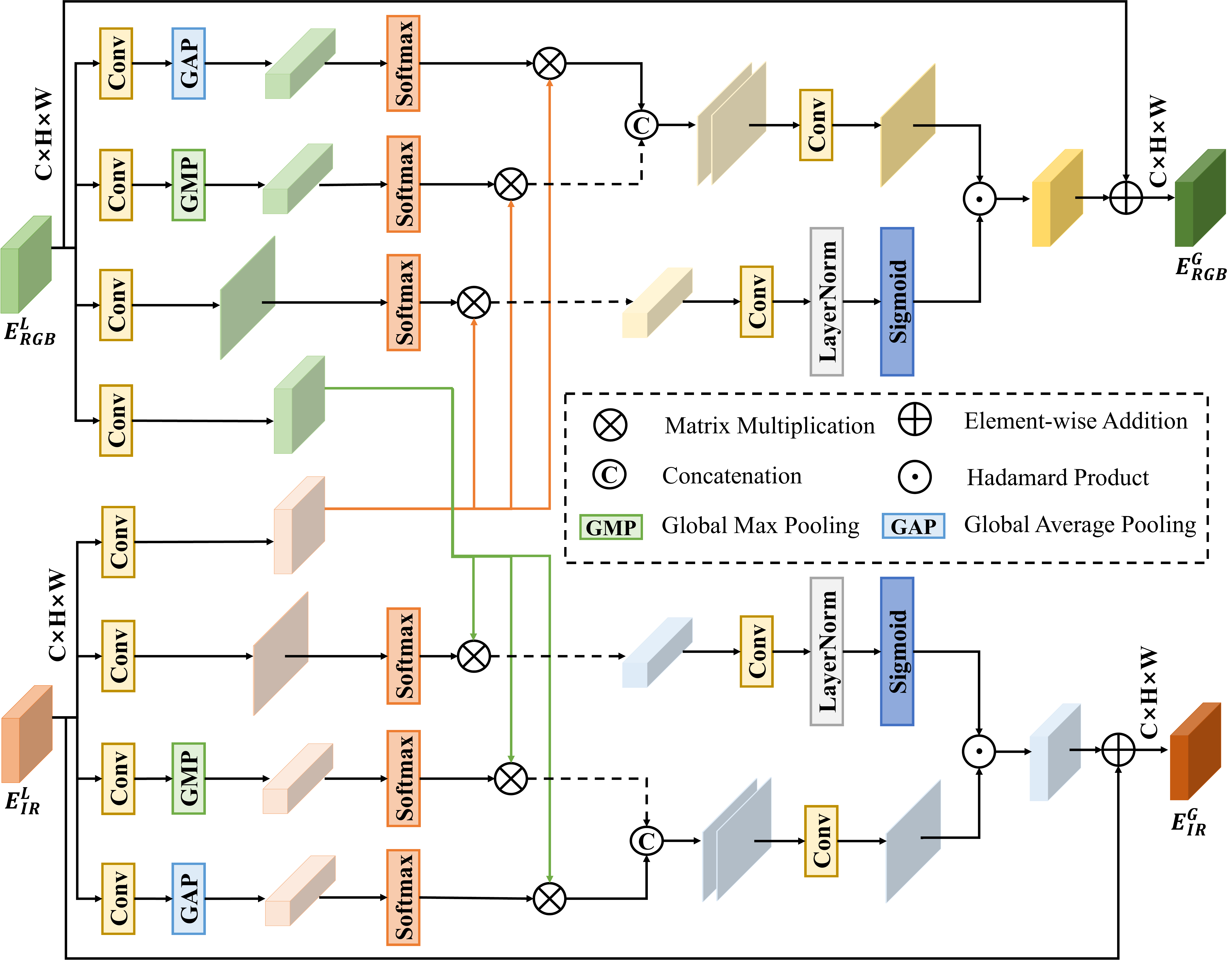}
		\caption{The structure of the designed cross-modal global modeling module (CGMM).}
        \label{fig4:CGMM}
	\end{center}
\end{figure}

Multimodal features have been enhanced with local contextual information by the LFEM. Subsequently, global modeling capability is crucial for removing inter-modality disparity and better understanding complex scenes, as it helps distinguish between objects and backgrounds and facilitates cross-modal feature fusion. Inspired by the CABlock \cite{Liu25} and SCAM \cite{Zhang24}, we proposed a cross-modal global modeling module (CGMM) to strengthen the global representation of cross-modal features, which can be displayed in Fig. \ref{fig4:CGMM}, where four parallel branches are used for each modality. The first two branches use the global average pooling (`GAP') and global max pooling (`GMP') to highlight spatial global information for cross-modal features, the third stream employs a \text{$1 \times 1$} convolution layer (`Conv') with the BN and SiLU to produce nonlinear channel results of multimodal feature maps, and the rest branch utilizes the same convolution operation with the normalization and activation function as the third branch to improve the representation capacity of inter-modality features. Later, the pixel-wise feature processing is adopted to promote information interaction and exchange to eliminate multimodal differences by using the matrix multiplication operation (`$\otimes$'), where the fourth branch of one modality performs the matrix multiplication operation with the first three branches of another modality respectively after three softmax operations to mutually interact between RGB-IR modalities and achieve spatial-channel contextual information representation. Further, the concatenated two branches using the concatenation operation (`$\copyright$') through a convolution layer (`Conv') and the third branch followed by a convolution layer (`Conv'), layer normalization (`LayerNorm'), and sigmoid function (`Sigmoid') execute the broadcast Hadamard product (`$\odot$') to enhance global feature relationship for RGB and IR modalities. To retain more feature details, the feature enhanced by the LFEM is added to the product result to generate the final result for each modality.

\subsection{Frequency domain feature aggregation module}
\begin{figure*}[htbp]
	\begin{center}
		\includegraphics[width=0.8\textwidth]{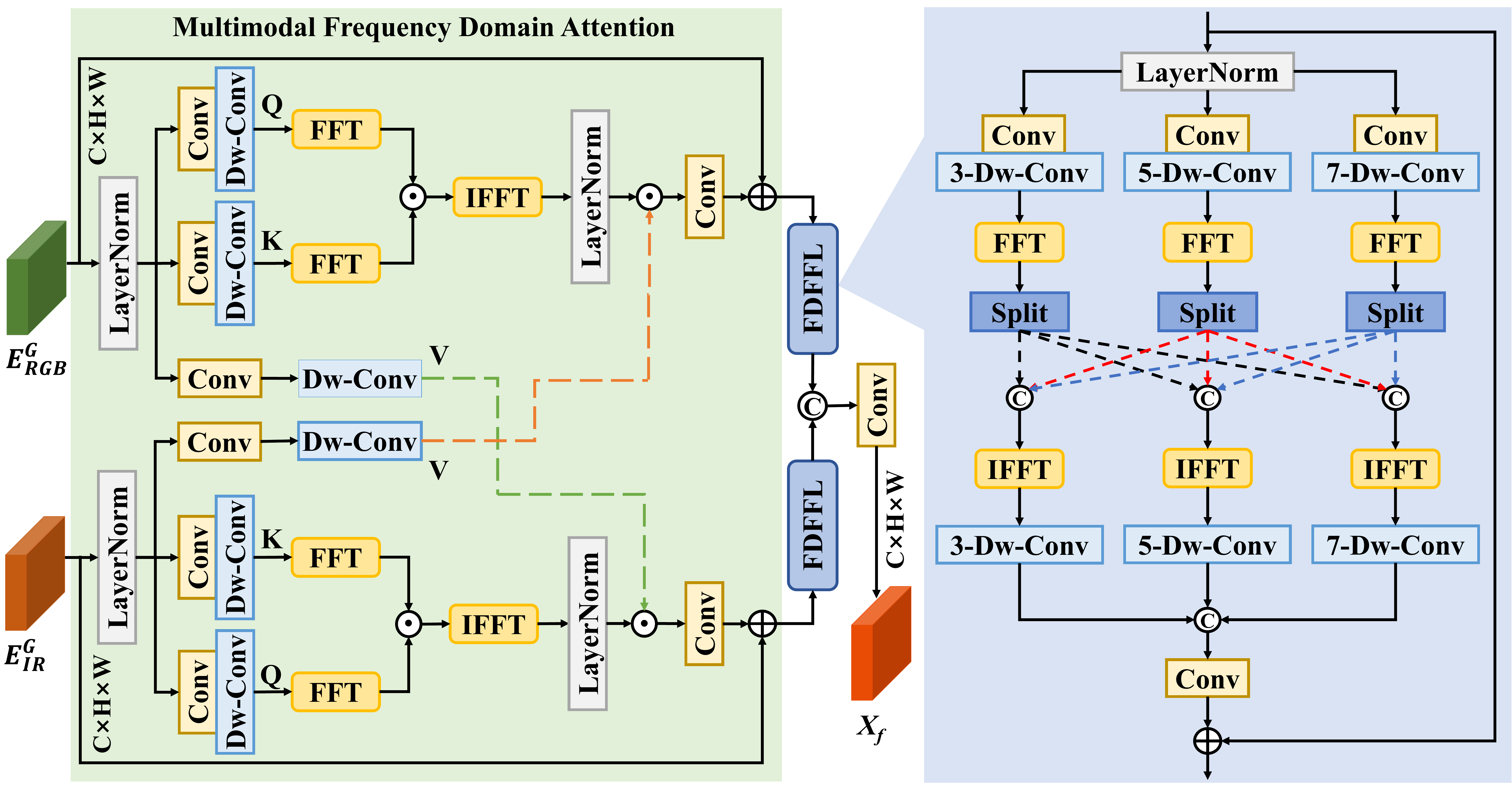}
		\caption{The structure of the designed frequency domain feature aggregation module (FDFAM). The FFT, IFFT, and FDFFL represents the fast fourier transform, inverse fast fourier transform, and frequency domain feed-forward layer, respectively.}
        \label{fig5:FDFAM}
	\end{center}
\end{figure*}

Compared to existing multimodal detection methods \cite{Hu25}, \cite{Yuan24}, \cite{Shen24} of using spatial domain transformers to extract complementary features from RGB-IR image pairs, we design a frequency domain feature aggregation module (FDFAM) to better exploit and fuse cross-modal complementary features, which consists of a multimodal frequency domain attention (MFDA), two frequency domain feed-forward layers (FDFFLs), a concatenation operation, and a convolution layer with an activation function, as shown in Fig. \ref{fig5:FDFAM}. The above detection methods use the cross attention mechanism with the matrix multiplication operation in the spatial domain to measure the similarity between tokens and capture complementary multimodal features. Inspired by the convolution theorem, which states that the correlations of two signals in the spatial domain are equivalent to their element-wise product in the frequency domain, we propose an effective MFDA to capture the pixel-wise correlations between cross-modal information. Initially, the query Q, key K, and value V are generated by the standard \text{$1 \times 1$} convolution layer and \text{$3 \times 3$} depth-wise convolution layer after a layer normalization (`LayerNorm') for each modality. Subsequently, the fast fourier transform (FFT) is introduced to execute domain transformation on the Q and K of RGB and IR features, and the element-wise multiplication (`$\odot$') is utilized to calculate the similarity between Q and K for different modalities. Later, we apply the inverse fast fourier transform (IFFT) and layer normalization to restore the frequency information to spatial feature and normalize it, respectively. To capture the inter-modality correlations, the V of one modality and the normalized result of another modality are interacted by the element-wise multiplication. Further, we employ the \text{$1 \times 1$} convolution layer and addition operation to obtain the output result of the proposed MFDA. The above procedure can be formulated as follows:
\begin{align}\tiny
    \begin{aligned}
    & Q_{RGB},K_{RGB},V_{RGB} =  f^{dwconv}_{3\times3}(f^{conv}_{1\times1}(LN(E_{RGB}^G))),\\
    & Q_{IR},K_{IR},V_{IR} =  f^{dwconv}_{3\times3}(f^{conv}_{1\times1}(LN(E_{IR}^G))),\\
    & F_{RGB}^{out} = E_{RGB}^G + f^{conv}_{1\times1}(LN(\overline{\mathcal{F}}(\mathcal{F}(Q_{RGB}) \odot \mathcal{F}(K_{RGB}))) \odot V_{IR}),\\
    & F_{IR}^{out} = E_{IR}^G + f^{conv}_{1\times1}(LN(\overline{\mathcal{F}}(\mathcal{F}(Q_{IR}) \odot \mathcal{F}(K_{IR}))) \odot V_{RGB}),
    \end{aligned}
\end{align}

where \text{$F_{rgb}^{out}$} and \text{$F_{ir}^{out}$} are the output of the MFDA. \text{$f^{conv}_{1\times1}$}, \text{$f^{dwconv}_{3\times3}$}, and LN stand for the standard \text{$1\times1$} convolution layer, \text{$3\times3$} depth-wise convolution layer, and layer normalization, respectively. \text{$\mathcal{F}$} and \text{$\overline{\mathcal{F}}$} represent the fast fourier transform and inverse fast fourier transform, separately.

Instead of the single-scale design, multi-scale representation is important for multispectral object detection \cite{Shen24}, which can exploit the details of objects with different sizes. Therefore, we devise a frequency domain feed-forward layer (FDFFL) via a mixed-scale frequency feature fusion strategy, which is displayed in Fig. \ref{fig5:FDFAM}. Through a layer normalization for each modality feature, we send the normalized feature \text{$X_l = LN(X)$} into three branches to capture multi-scale representation with three standard convolution layers and three depth-wise convolution layers of different kernels, and the ReLU activation function follows the depth-wise convolution layer to increase the nonlinearity. Then, we convert spatial feature information into frequency information through the FFT operation. To enhance multi-scale information for frequency feature, we split each frequency information into three chunks in the channel dimension and concatenate these chunks in a mutual mixing manner. The above processes can be defined as follows:
\begin{align}
    \begin{aligned}
    & X_3^{c1}, X_3^{c2}, X_3^{c3} = \varsigma(\mathcal{F}(\sigma(f^{dwconv}_{3\times3}(f^{conv}_{1\times1}(X_l))))), \\
    & X_5^{c1}, X_5^{c2}, X_5^{c3} = \varsigma(\mathcal{F}(\sigma(f^{dwconv}_{5\times5}(f^{conv}_{1\times1}(X_l))))), \\
    & X_7^{c1}, X_7^{c2}, X_7^{c3} = \varsigma(\mathcal{F}(\sigma(f^{dwconv}_{7\times7}(f^{conv}_{1\times1}(X_l))))), \\
    \end{aligned}
\end{align}
where \text{$\sigma$} and \text{$\varsigma$} represent the ReLU activation function and split operation, respectively. \text{$f^{conv}_{1\times1}$}, \text{$f^{dwconv}_{3\times3}$}, \text{$f^{dwconv}_{5\times5}$}, and \text{$f^{dwconv}_{7\times7}$} denote the \text{$1 \times 1$} standard convolution layer, \text{$3 \times 3$}, \text{$5 \times 5$}, and \text{$7 \times 7$} depth-wise convolution layers, respectively. Subsequently, we restore the merged result to spatial information through the IFFT operation. These mixed features are concatenated by a concatenation operation following different depth-wise convolution layers and ReLU activation function, and the concatenated feature is reduced in dimension using \text{$1\times1$} convolution layer and added to the initial information \text{$X$} to obtain the final result of the FDFFL. The process is formulated as:
\begin{align}
    \begin{aligned}
    & X_3 = \sigma(f^{dwconv}_{3\times3}(\overline{\mathcal{F}}([X_3^{c1},X_5^{c1}, X_7^{c1}]))), \\
    & X_5 = \sigma(f^{dwconv}_{5\times5}(\overline{\mathcal{F}}([X_3^{c2},X_5^{c2}, X_7^{c2}]))), \\
    & X_7 = \sigma(f^{dwconv}_{7\times7}(\overline{\mathcal{F}}([X_3^{c3},X_5^{c3}, X_7^{c3}]))), \\
    & \bar{X} = X + f^{conv}_{1\times1}([X_3, X_5, X_7])
    \end{aligned}
\end{align}
where \text{$\bar{X}$} is the output of the FDFFL, and [, ,] represent the concatenation operation.

After the RGB and IR features are enhanced by the multi-scale representation of the FDFFL, we use a concatenation operation and a \text{$1 \times 1$} convolution layer with a ReLU activation function to obtain the fused feature \text{$X_f$}, which can be expressed as follows:
\begin{align}
    \begin{aligned}
        X_f = \sigma(f^{conv}_{1\times1}([X_{RGB}, X_{IR}])),
    \end{aligned}
\end{align}

where \text{$X_{rgb}$} and \text{$X_{ir}$} are the outputs of the FDFFL for different modalities. \text{$\sigma$}, \text{$f^{conv}_{1\times1}$} , and [,] stand for the ReLU activation function, \text{$1 \times 1$} standard convolution layer, and concatenation operation, respectively.

\subsection{Loss function}
The loss function of our proposed FreDFT embedded into the YOLOv5 detector is composed of the regression loss ($\mathcal{L}_{box}$), the classification loss ($\mathcal{L}_{cls}$), and the confidence loss ($\mathcal{L}_{obj}$), which is defined as:
\begin{equation}
    \mathcal{L} = \mathcal{L}_{box} + \mathcal{L}_{cls} + \mathcal{L}_{obj},
\end{equation}
where $\mathcal{L}_{box}$ uses the complete intersection over union (CIoU) loss \cite{Zheng20}. $\mathcal{L}_{cls}$ and $\mathcal{L}_{obj}$ adopt the cross-entropy (CE) loss. In order to solve the problem of category imbalance, following \cite{Guo24}, the varifocal loss \cite{ZhangW21} is introduced to dynamically adjust the weight of the CE loss and improve the discrimination ability of rare classes.

\section{EXPERIMENTS}\label{Sec_Experiments}
\subsection{Datasets}
We trained and tested the proposed FreDFT using three public datasets: FLIR, LLVIP, and M$^3$FD.
\subsubsection{FLIR}
The FLIR dataset \cite{Zhang20} is potentially challenging for multispectral object detection task, which consists of 5142 visible-infrared image pairs with a resolution of \text{$640 \times 520$} captured in day and night traffic environments, where 4129 image pairs are used for model training, and 1013 image pairs are applied for model testing. This dataset contains three categories: ``person", ``car", and ``bicycle".

\subsubsection{LLVIP}
The LLVIP dataset \cite{Jia21} is a large-scale multispectral pedestrian dataset collected in low-light conditions, which consists of 15488 well-aligned visible-infrared image pairs with a resolution of \text{$1280 \times 1024$}, where 12025 image pairs are utilized for model training, and 3463 image pairs are used to test the model performance. This dataset includes one category: ``person".

\subsubsection{M$^3$FD}
The M$^3$FD dataset \cite{Liu22} is a slight misalignment benchmark dataset captured in scenes such as complex traffic and occluded environments, which consists of 4200 visible-infrared image pairs with a resolution of \text{$1024 \times 768$}. Following \cite{Liang23}, 3360 and 840 image pairs are employed for model training and evaluation, respectively. This dataset consists of six categories: ``People", ``Car", ``Bus", ``Lamp", ``Motorcycle", and ``Truck".

\subsection{Implementation details}
We run the proposed FreDFT method with a batch size of 2 on a single NVIDIA GeForce RTX 3090 GPU under the PyTorch framework. The SGD optimizer is used to update the model parameters, where the initial learning rate (lr) $1.0 \times 10^{-2}$, momentum 0.937, and weight decay $5.0 \times 10^{-4}$  are set. Also, the warmup and cosine annealing strategy is used for the training process of our FreDFT model. Data augmentation methods such as random rotation and mosaic are applied to improve the model robustness. We set the training epoch to 150, 100, and 200 for the FLIR, LLVIP, and M$^3$FD datasets, respectively.

\subsection{Comparison with state-of-the-art methods}
Our proposed FreDFT is compared with other state-of-the-art detection methods on multiple public datasets, including FLIR, LLVIP, and M$^3$FD. In order to show the advantages of the FreDFT in more detail, we provide quantitative analysis and visual detection effect experiments.

\subsubsection{Evaluation on the FLIR Dataset}
\begin{table}[htbp]
\renewcommand\arraystretch{1.5}
\centering
\caption{The detection results between our proposed FreDFT and other methods on the FLIR dataset. The best results are marked in bold.}
\label{Tab1_FLIR}
\begin{tabular}{cccc}
\hline
Models & Modality Type & mAP50($\%$) & mAP($\%$)\\
\hline
Faster R-CNN \cite{Ren17} &  RGB & 65.0 & 30.2\\

Faster R-CNN \cite{Ren17}  & IR & 73.4  & 37.9\\

YOLOv5 \cite{YOLOv5} & RGB & 67.8 & 31.2\\

YOLOv5 \cite{YOLOv5} & IR & 74.4 & 38.0\\
\hline
YOLOFusion \cite{Fang22} & RGB+IR & 76.6 & 39.8\\

CSAA \cite{Cao23}  & RGB+IR  & 79.2 & 41.3 \\

CFT \cite{Fang21} & RGB+IR & 78.3 & 40.2\\

YOLO-MS \cite{Xie23} & RGB+IR & 75.3 & 38.3\\

GAFF \cite{Zhang21} & RGB+IR & 72.9 & 37.5\\

ProbEn \cite{Chen22} & RGB+IR & 75.5 & 37.9\\

MSANet \cite{You23} & RGB+IR & 76.2 & 39.0\\

MMFN \cite{Yang24} & RGB+IR & 80.8 & 41.7\\

SeaDate \cite{Dong24} & RGB+IR & 80.3 & 41.3\\

CrossFormer \cite{Lee24} & RGB+IR  & 79.3 & 42.1 \\

ICAFusion \cite{Shen24} & RGB+IR & 79.2 & 41.4\\

MMI-Det \cite{Zeng24} & RGB+IR & 79.8 & 40.5\\

EI$^2$Det \cite{HuH25} & RGB+IR & 80.2 & -\\
\hline
FreDFT & RGB+IR & $\mathbf{83.5}$ & $\mathbf{42.6}$\\
\hline
\end{tabular}
\end{table}

\begin{table}[htbp]
\renewcommand\arraystretch{1.5}
\centering
\caption{The mAP50 and mAP values of our FreDFT and other approaches on the LLVIP dataset. The best results are highlighted in bold.}
\label{Tab2_LLVIP}
\begin{tabular}{cccc}
\hline
Models & Modality Type & mAP50($\%$) & mAP($\%$)\\
\hline
Faster R-CNN \cite{Ren17} & RGB & 88.8 & 47.5 \\

Faster R-CNN \cite{Ren17} & IR & 92.6 & 50.7\\

YOLOv5 \cite{YOLOv5} & RGB & 90.8 & 50.0\\

YOLOv5 \cite{YOLOv5} & IR & 94.6 & 61.9\\
\hline
YOLOFusion \cite{Fang22} & RGB+IR & 95.4 & 62.8\\

TINet \cite{ZhangY23}  & RGB+IR & 94.3 & 60.6\\

TarDAL \cite{Liu22} & RGB+IR & 94.9 & 61.2\\

RSDet \cite{Zhao24} & RGB+IR & 95.8 & 61.3\\

CSAA \cite{Cao23} & RGB+IR & 94.3 & 59.2\\

CFT \cite{Fang21} & RGB+IR & 97.5 & 63.6\\

GAFF \cite{Zhang21} & RGB+IR & 94.0 & 55.8\\

ProbEn \cite{Chen22} & RGB+IR & 93.4 & 51.5\\

CCIFNet \cite{Yan23} & RGB+IR & 97.6 & 64.1\\

CrossFormer \cite{Lee24} & RGB+IR  & 97.5 & 65.1\\

MMFN \cite{Yang24} & RGB+IR & 97.2 & -\\

ICAFusion \cite{Shen24} & RGB+IR & 96.3 & 62.3\\
\hline
FreDFT & RGB+IR & $\mathbf{98.2}$ & $\mathbf{66.0}$ \\
\hline
\end{tabular}
\end{table}

\begin{figure*}[htbp]
	\begin{center}
		\includegraphics[width=0.75\textwidth]{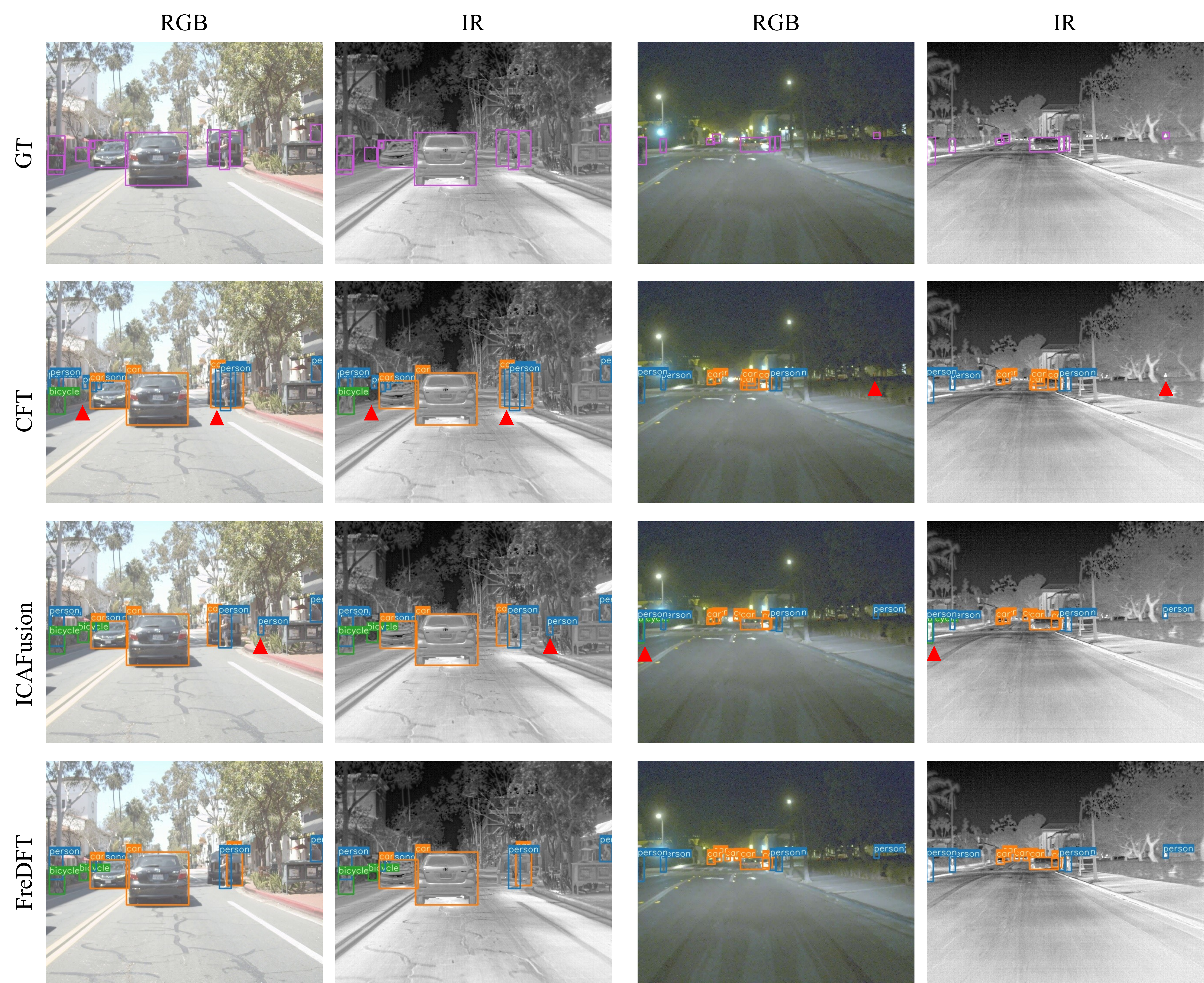}
		\caption{The visual prediction results of the ground truth (GT), CFT, ICAFusion, and our FreDFT for day (left image pair) and night (right image pair) scenes in the FLIR dataset, where the red triangles mark wrong prediction bounding boxes.}
        \label{fig6:FLIR}
	\end{center}
\end{figure*}

In Table \ref{Tab1_FLIR}, we conduct a quantitative comparison between our proposed FreDFT and many state-of-the-art methods on the FLIR dataset. It can be seen that the FreDFT outperforms other methods on the mAP50 and mAP metrics, which demonstrates the effectiveness and superiority of our method in both daytime and nighttime traffic scenarios. Moreover, we run a visual comparison of the detection effects on two RGB and IR image pairs of day and night from the FLIR dataset, as shown in Fig. \ref{fig6:FLIR}. We can find that the CFT method misses the detection objects of the ``people" and ``bicycle" categories, and the state-of-the-art ICAFusion method generates incorrect detection results for these two categories. These prediction results are marked by the red triangles in Fig. \ref{fig6:FLIR}. Compared with the CFT and ICAFusion, our proposed FreDFT achieves more accurate detection and has better localization capabilities.

\subsubsection{Evaluation on the LLVIP Dataset}
We report a numerical evaluation in Table \ref{Tab2_LLVIP}, where our designed FreDFT and other state-of-the-art approaches are compared on the LLVIP dataset. We can conclude from Table \ref{Tab2_LLVIP} that the FreDFT achieves the best detection performance on both mAP50 and mAP evaluation indicators. Specifically, compared with the best detection performance of monomodal detectors, our FreDFT outperforms them by 3.4\% and 4.1\% on mAP50 and mAP values, respectively. Also, other excellent multimodal detection models, such as CrossFormer \cite{Lee24}, MMFN \cite{Yang24}, and ICAFusion \cite{Shen24}, are inferior to our FreDFT.

\subsubsection{Evaluation on the M$^3$FD Dataset}
\begin{table}[htbp]
\renewcommand\arraystretch{1.5}
\centering
\caption{Comparison of detection results between our designed FreDFT and other models on the M$^3$FD dataset. The top results are emphasized in bold.}
\label{Tab3_M3FD}
\begin{tabular}{cccc}
\hline
Models & Modality Type & mAP50($\%$) & mAP($\%$)\\
\hline
YOLOv5 \cite{YOLOv5} & RGB & 73.8 & 42.9\\

YOLOv5 \cite{YOLOv5} & IR & 68.6 & 40.4\\

YOLOv8 \cite{YOLOv8} & RGB & 80.9 & 52.5\\

YOLOv8 \cite{YOLOv8} & IR & 79.5 & 53.1\\
\hline
DIDFuse \cite{Zhao20} & RGB+IR & 78.9 & 52.6\\

SDNet \cite{ZhangM21} & RGB+IR & 79.0 & 52.9\\

CFT \cite{Fang21} & RGB+IR & 88.2 & 59.0 \\

RFNet \cite{Xu22} & RGB+IR & 79.4 & 53.2\\

SuperFusion \cite{TangD22} & RGB+IR & 83.5 & 56.0\\

DeFusion \cite{SunC22} & RGB+IR & 80.8 & 53.8\\

TarDAL \cite{Liu22} & RGB+IR & 80.5 & 54.1\\

IGNet \cite{Li23} & RGB+IR & 81.5 & 54.5\\

CDDFuse \cite{ZhaoB23} & RGB+IR & 81.1 & 54.3\\

ICAFusion \cite{Shen24} & RGB+IR & 87.1 & 57.0 \\
\hline
FreDFT & RGB+IR & $\mathbf{88.4}$ & $\mathbf{59.7}$\\
\hline
\end{tabular}
\end{table}

\begin{figure*}[htbp]
	\begin{center}
		\includegraphics[width=0.70\textwidth]{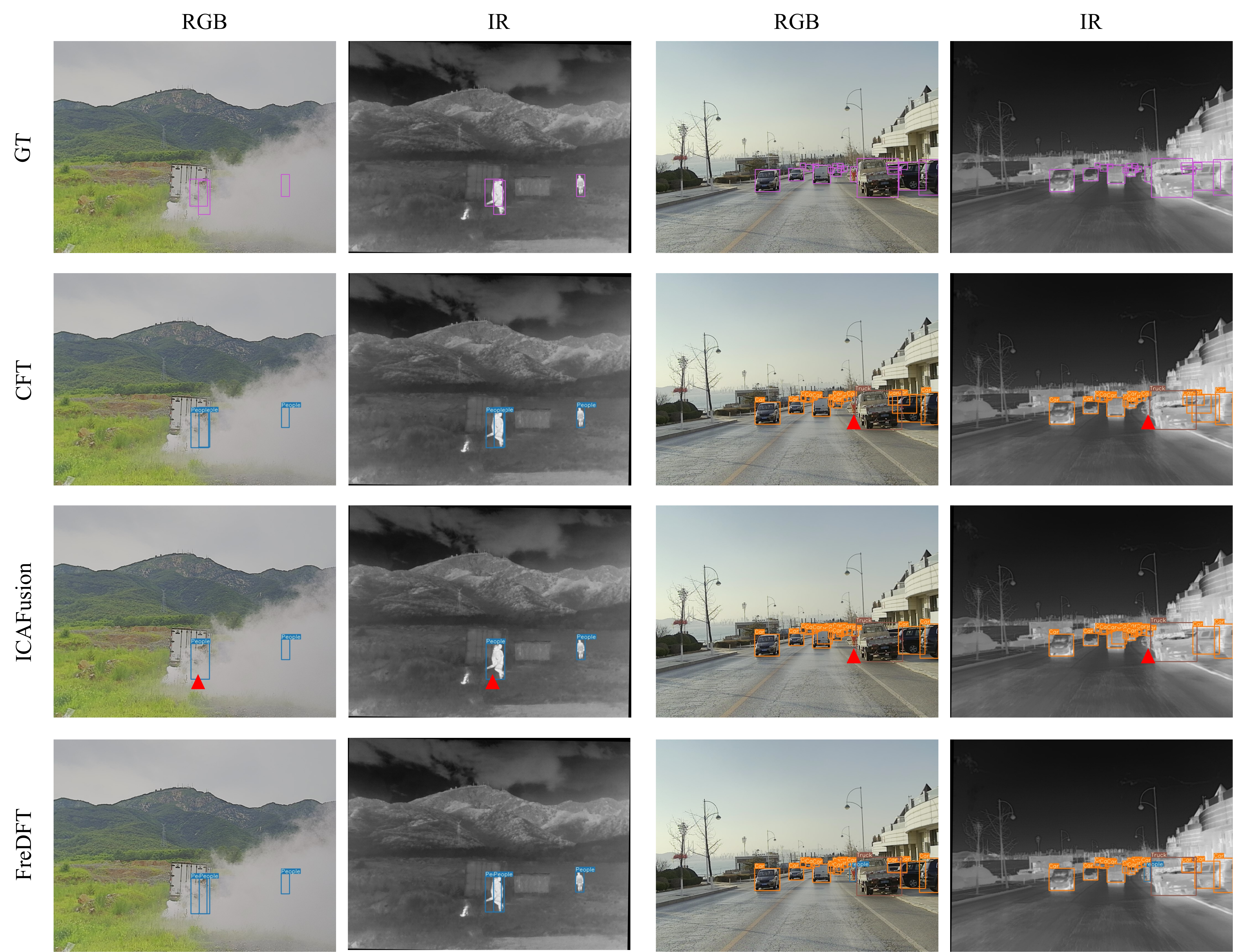}
		\caption{The two detection samples of the ground truth (GT), CFT, ICAFusion, and our FreDFT for thick smoke (left image pair) and complex traffic (right image pair) scenes on the M$^3$FD dataset, where the red triangles denote incorrect prediction bounding boxes.}
        \label{fig7:M3FD}
	\end{center}
\end{figure*}

\begin{figure*}[htbp]
	\begin{center}
		\includegraphics[width=0.70\textwidth]{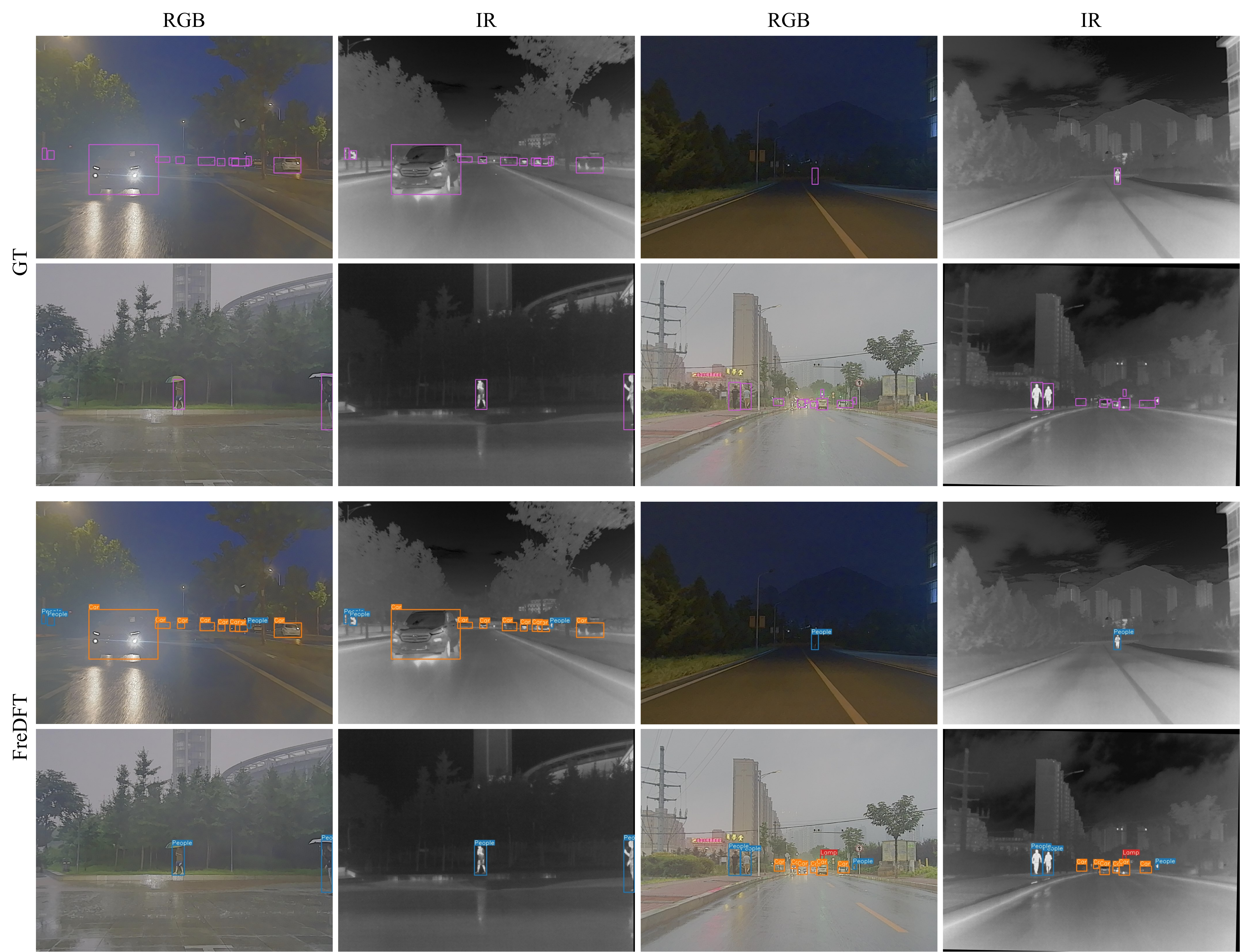}
		\caption{The four detection samples of the ground truth (GT) and our FreDFT for low-light, rain, and fog conditions on the M$^3$FD dataset.}
        \label{fig7:M3FD_1}
	\end{center}
\end{figure*}

The quantitative comparison results are displayed in Table \ref{Tab3_M3FD}, where our proposed FreDFT is compared with other state-of-the-art models on the M$^3$FD dataset. It can be found that the FreDFT is superior to other methods on the mAP50 and mAP values, which suggests that the FreDFT provides a more effective solution for visible and infrared images. Further, we visualize the prediction results to directly reflect the detection performance of different methods, as shown in Fig. \ref{fig7:M3FD}. We can see that in the case of heavy smoke and partial occlusion, the state-of-the-art ICAFusion cannot accurately identify the ``people" category, but our proposed FreDFT method can identify this object. Moreover, in complex traffic scenes, our FreDFT generates precise prediction bounding boxes, while the CFT and ICAFusion miss the ``people" class in the detected area, which are marked by the red triangles in Fig. \ref{fig7:M3FD}. Also, Fig. \ref{fig7:M3FD_1} illustrates that our FreDFT can accurately predict objects in different environments, such as low-light, rain, and fog.

\subsection{Complexity analysis}
\begin{table}[htbp]
\renewcommand\arraystretch{1.5}
\centering
\caption{The model complexity of different detection methods, where the multimodal image resolution is adjusted to $640\times640$.}
\label{Tab4_complexity}
\begin{tabular}{cccc}
\hline
Models & Modality Type & Params(M) & FLOPs(G) \\
\hline
CFT \cite{Fang21} & RGB+IR & 206.0 & 224.6 \\

ProbEn \cite{Chen22} & RGB+IR & 945.9 & - \\

MMFN \cite{Yang24} & RGB+IR & 176.4 & - \\

SeaDate \cite{Dong24} & RGB+IR & 306.6 & 134.3 \\

CrossFormer \cite{Lee24} & RGB+IR & 340.0 & 361.7\\

ICAFusion \cite{Shen24} & RGB+IR & 120.2 & 192.7 \\

EI$^2$Det \cite{HuH25} & RGB+IR & 127.7 & 220.9\\

MMI-Det \cite{Zeng24} & RGB+IR & 207.6 & 229.2\\

FreDFT & RGB+IR & 152.6 & 464.5 \\
\hline
\end{tabular}
\end{table}

To fully evaluate the proposed FreDFT, we conduct a comparative experiment on model parameters and floating-point operations per second (FLOPs), which can be listed in \ref{Tab4_complexity}, where we select many state-of-the-art methods, including the CFT \cite{Fang21}, ProbEn \cite{Chen22}, MMFN \cite{Yang24}, SeaDate \cite{Dong24}, CrossFormer \cite{Lee24}, ICAFusion \cite{Shen24}, EI$^2$Det \cite{HuH25}, and MMI-Det \cite{Zeng24}, to make a comparison with our FreDFT. We can see that the number of parameters of our FreDFT model is less than that of the CFT, ProbEn, MMFN, SeaDate, CrossFormer, and MMI-Det, but more than that of the ICAFusion and EI$^2$Det. Although the FLOPs of our FreDFT is higher than other compared models, the detection accuracy of the FreDFT is higher than other detection methods.

\begin{figure}[htbp]
	\begin{center}
		\includegraphics[width=0.43\textwidth]{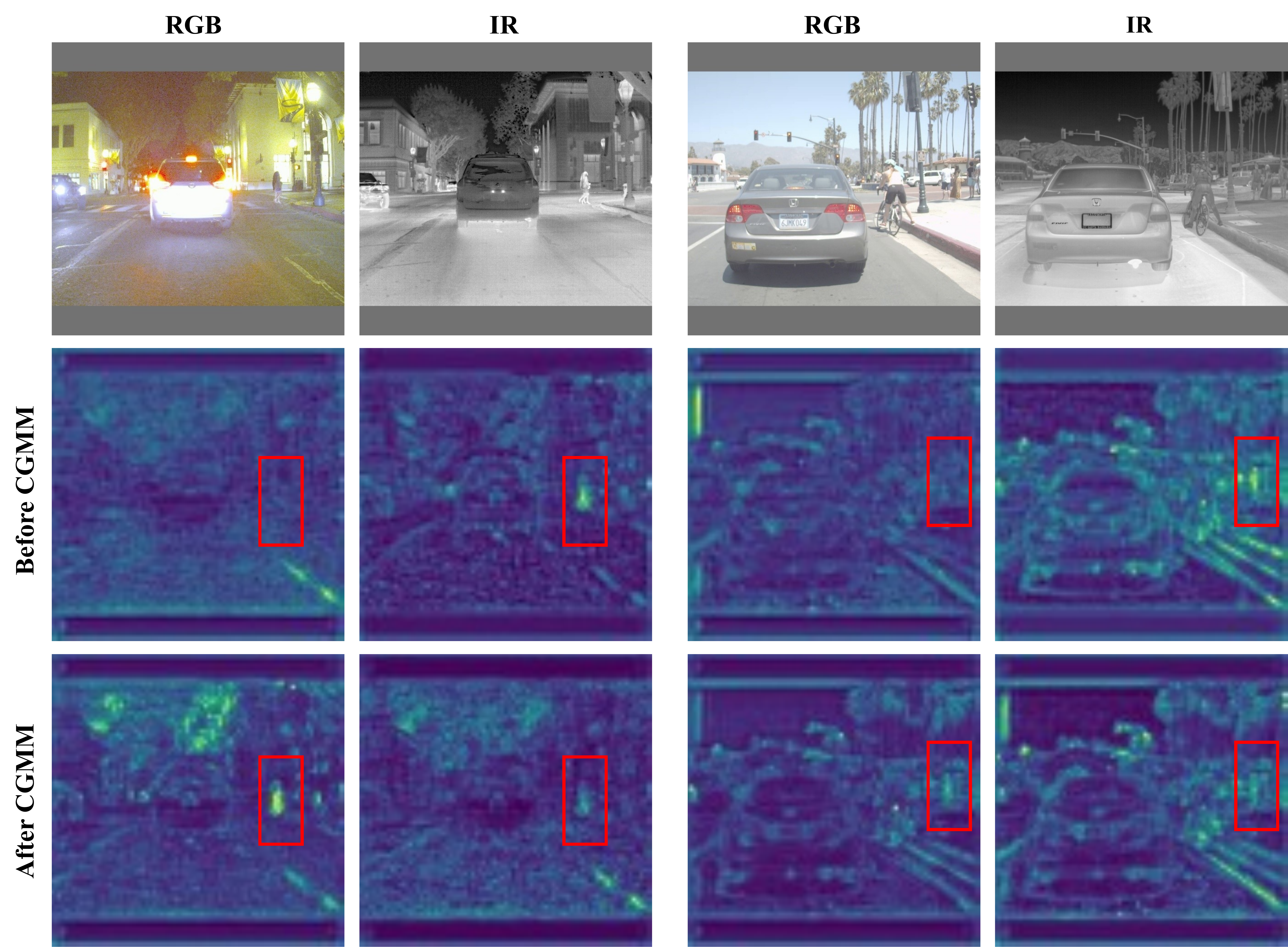}
		\caption{Feature map visualization before and after CGMM module, where the size of the feature maps is $80 \times 80$. Two pairs of day and night RGB-IR images are selected from the FLIR dataset.}
        \label{fig8:Feature}
	\end{center}
\end{figure}

\begin{figure*}[htbp]
	\begin{center}
		\includegraphics[width=\textwidth]{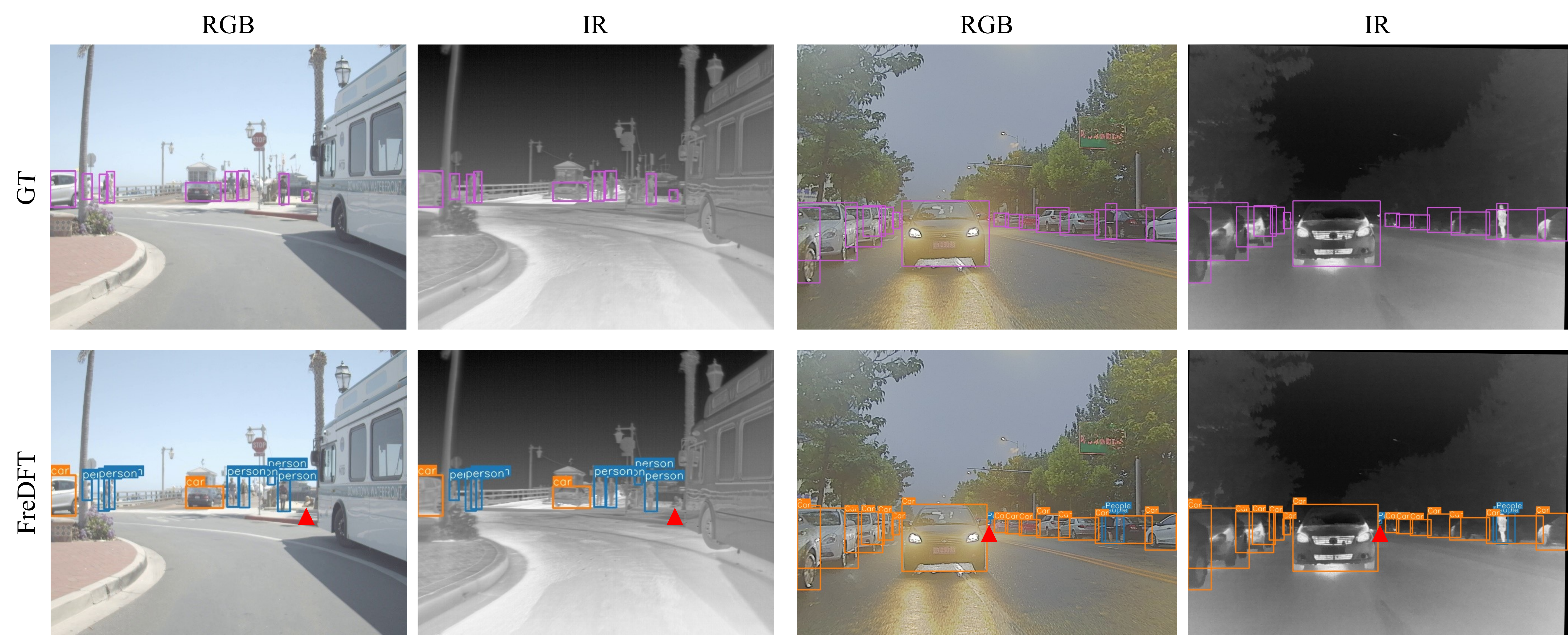}
		\caption{Failure detection cases on two RGB-IR image pairs from the FLIR and M$^3$FD datasets, respectively.}
        \label{fig9:Failure}
	\end{center}
\end{figure*}

\subsection{Ablation study}
\begin{table}[htbp]
\renewcommand\arraystretch{1.5}
\centering
\caption{The impact of the LFEM, CGMM and FDFAM in our proposed FreDFT on the model performance, where the FLIR dataset is used for evaluation.
The LFEM, CGMM and FDFAM represent local feature enhancement module, cross-modal global modeling module and frequency domain feature aggregation module, respectively.}
\label{Tab5_Module_ablation}
\begin{tabular}{ccccc}
\hline
LFEM & CGMM & FDFAM & mAP50($\%$) & mAP($\%$)\\
\hline
$\times$ & $\times$ & $\times$ & 79.8 & 41.7\\

$\times$ & $\times$ & $\checkmark$ & 81.7 & 42.3\\

$\checkmark$ & $\times$ & $\checkmark$ & 82.8 & 42.4\\

$\times$ & $\checkmark$ & $\checkmark$ & 82.2 & 42.2\\

$\checkmark$ & $\checkmark$ & $\checkmark$ & 83.5 & 42.6\\
\hline
\end{tabular}
\end{table}

\begin{table}[htbp]
\renewcommand\arraystretch{1.5}
\centering
\caption{The detection performance comparison of different cross attention on the FLIR dataset, where the MSDA and MFDA denote multimodal spatial domain attention and multimodal frequency domain attention, separately.}
\label{Tab6_CA_ablation}
\begin{tabular}{ccc}
\hline
  & mAP50($\%$) & mAP($\%$) \\
\hline
MSDA & 82.4 & 42.2 \\

MFDA & 83.5 & 42.6 \\
\hline
\end{tabular}
\end{table}

\begin{table}[htbp]
\renewcommand\arraystretch{1.5}
\centering
\caption{The performance comparison of two feed-forward networks on the M$^3$FD dataset, where the MLP is the multilayer perceptron, and the FDFFL denotes our proposed frequency domain feed-forward layer.}
\label{Tab8_MLP_ablation}
\begin{tabular}{cccc}
\hline
  & mAP50($\%$) & mAP($\%$) \\
\hline
standard MLP & 87.6 & 59.3 \\

FDFFL & 88.4 & 59.7 \\
\hline
\end{tabular}
\end{table}

To better assess the effectiveness of each module in the proposed FreDFT, we execute an ablation experiment in Table \ref{Tab5_Module_ablation} to judge the performance impact of the designed local feature enhancement module (LFEM), cross-modal global modeling module (CGMM), and frequency domain feature aggregation module (FDFAM) for the proposed method. Following the CFT \cite{Fang21}, SeaDate \cite{Dong24} and MMFN \cite{Yang24}, we use simple addition operation to fuse RGB and IR features extracted by dual backbone network as the baseline model to test the detection performance without any our designed modules, and the results are listed in the second row of Table \ref{Tab5_Module_ablation}. Subsequently, we apply the proposed FDFAM to fuse multimodal features to generate the detection results, and then continue to add the LFEM and CGMM before cross-modal feature fusion to obtain the prediction results. From Table \ref{Tab5_Module_ablation}, we can find that the proposed FreDFT containing the LFEM, CGMM, and FDFAM achieves the best detection precision, which indicates that these modules are crucial and indispensable for the detection results. Also, we performed a visual comparison experiment on the ability of the CGMM module to alleviate inter-modality differences, as shown in Fig. \ref{fig8:Feature}. Specifically, it can be clearly seen from the red boxes in Fig. \ref{fig8:Feature} that the CGMM can reduce the information imbalance between modalities, which proves that the CGMM can eliminate inter-modal conflicts by performing cross-modal information interaction in a spatial and channel manner.

Additionally, we analyze the impact of different cross-attention and feed-forward networks on the model detection performance. In Table \ref{Tab6_CA_ablation}, we compared the mAP values of multimodal spatial domain attention (MSDA) and our proposed multimodal frequency domain attention (MFDA). It can be seen that the MFDA is superior to the MSDA in terms of mAP50 and mAP metrics, which indicates that our proposed MFDA can achieve higher quality detection and more accurate localization. Further, in the MSDA, the correlations between multimodalities are obtained by the matrix multiplication operation, which has a significantly higher computational complexity than element-wise multiplication in the MFDA, where the computational cost of domain transformation and softmax operation are ignored. Besides, compared with the FLIR dataset, the M$^3$FD dataset contains more categories with different sizes, which can more comprehensively test the performance of our proposed frequency domain feed-forward layer (FDFFL) and the standard multilayer perceptron (MLP). So we perform a comparison experiment on the M$^3$FD dataset, which is displayed in Table \ref{Tab8_MLP_ablation}. We can find that the FDFFL outperforms vanilla MLP on mAP50 and mAP values, which illustrates the advantages of our FDFFL.

\section{Disscussion}\label{Sec_Disscussion}
To visually demonstrate the limitations of our FreDFT, we conduct a comparative experiment between our FreDFT and Ground Truth (GT) in Fig. \ref{fig9:Failure}, in which failed or incorrect detection results are marked with red triangles. In the first pair of RGB-IR images, we can find a missed detection box due to an occluded seated person whose behavior differed from that of most people walking on the roadside, resulting in feature discrepancies. Furthermore, the person is heavily occluded and a small object. The second image pair is taken in a low-light environment, and our FreDFT generates a wrong prediction box for a small object. The incorrect detection box area is blurry in the RGB image, but appears as a bright area in the IR image, causing our FreDFT to mistakenly identify a person. These challenges lead to the detection failure of our FreDFT.

To address these issues, occlusion simulation should be implemented to enable object detection performance using local features, and more pedestrian data, including non-upright postures such as sitting and squatting, should be collected. Furthermore, more advanced backbone networks and detection heads should be considered to improve detection results.

\section{CONCLUSION}\label{Sec_Conclusion}
In this work, we design a novel FreDFT architecture for visible-infrared object detection, which consists of a local feature enhancement module (LFEM), a cross-modal global modeling module (CGMM), and a frequency domain feature aggregate module (FDFAM). Firstly, we design a new LFEM to improve the local information representation capabilities between modalities, which uses different types of convolutional layers to focus on different local areas and applies a channel mixing strategy to enhance feature recombination at the channel level. Then, we develop a novel CGMM to eliminate cross-modal feature conflicts and provide beneficial conditions for multimodal fusion, which adopts global pooling to obtain global context information to construct latent relationships between modalities by pixel-level cross-modal spatial-channel interaction. Furthermore, we propose an effective FDFAM, which mainly consists of a multimodal frequency domain attention (MFDA) and a frequency domain feed-forward layer (FDFFL), to capture the pixel-by-pixel correlations between cross-modal information and achieve high-quality cross-modal feature fusion. Inspired by the convolution theorem, a MFDA is used instead of a multimodal spatial domain attention to capture cross-modal complementary features more effectively, and a FDFFL via a mixed-scale frequency feature fusion strategy is designed to promote cross-modal feature fusion. Comprehensive experiments demonstrate that our proposed FreDFT can generate excellent detection results compared with other state-of-the-art methods.

\ifCLASSOPTIONcaptionsoff
  \newpage
\fi

\bibliographystyle{IEEEtran}
\bibliography{reference}

\end{document}